\documentclass{jdmdh}
\usepackage{array}
\usepackage{pgfplots}

\usepackage[utf8]{inputenc}
\usepackage{geometry}
\usepackage{graphicx}
\usepackage{appendix}
\usepackage{latexsym}
\usepackage{multirow}
\usepackage{booktabs}

\pgfplotsset{compat=1.18}

\titlespacing*{\section}
{0pt}{2.5ex plus 1ex minus .2ex}{1.3ex plus .2ex}
\titlespacing*{\subsection}
{0pt}{1.0ex plus 1ex minus .2ex}{1.3ex plus .2ex}
\titlespacing*{\subsubsection}
{0pt}{1.0ex plus 1ex minus .2ex}{1.3ex plus .2ex}

\title{The Impact of Incumbent/Opposition Status and Ideological Similitude on Emotions in Political Manifestos}

\author[1]{Takumi Nishi}

\affil[1]{Waseda University, Japan} 

\corrauthor{Takumi Nishi}{t-nishi@fuji.waseda.jp/taknishi95@gmail.com}

\begin{document}

\maketitle
\abstract{
This study analyzes emotion-associated language in the United Kingdom's Conservative and Labour Parties' general election manifestos from 2001 to 2019. While previous research has shown a correlation between ideology and positions in policy, there are still conflicting results in matters of the sentiments in such manifestos. Using new data, we present how a manifestos' valence level can be swayed by a party's status in government, with incumbent parties presenting a higher frequency in positive emotion-associated words while negative emotion-associated words are more prevalent in opposition parties. We also demonstrate that parties with ideological similitude use positive language prominently, further adding to the literature on the relationship between sentiments and party status.}

\section{Introduction}
In an age where social media and other digital communication tools are key mediums for mass communications, various political parties have dedicated their effort to establishing their social media strategy and capitalizing on its strengths \citep{yildirim_2020}. And consequently, there has been a growing perception within democratic societies viewing political manifestos as an obsolete relic \citep{bogdanor_2015}. While it can be suggested that the impact of manifestos has relatively waned over the decades, many scholars continue to emphasize their importance in addressing various parties' campaign strategies \citep{manifesto_important}. In particular, manifestos enable parties to differentiate their policy from their competitors while streamlining the campaign strategy for its politicians. Additionally, recent studies on manifestos indicate the importance of emotive language in the text as a crucial asset in persuading voters \citep{manifesto_important, ohman_fin}. In effect, we offer that political manifestos can provide crucial data to uncover a political party's strategy in deploying emotions, especially during elections.

While the recent interest in social media's role in politics has produced many studies in sentiment analysis of social media posts such as tweets, there is little research on the same topic using manifesto texts \citep{social-media-research, ohman_fin}. Using 12 general election manifestos from the United Kingdom's two establishment parties, the Conservative Party and the Labour Party, published between 2001 and 2019 as data, we will utilize a lexicon-based emotional analysis approach to investigate how a party's position in government (incumbent or opposition), alongside, ideological similarities affect the use and frequency of emotive languages in manifestos over time. 

\section{Previous Research and Hypotheses}
At the current level, previous manifesto research related to sentiment/emotion analysis revolves around various European parities' campaign materials \citep{crab_2018, topic_german,ohman_fin}, of which many show evidence that the party's status in government and ideological positioning have a standing influence on these issues. Based on these findings, the study establishes two hypotheses (H) as follows:

\textit{\textbf{Hypothesis 1 (H1):  The frequency of positive emotions will be highest in the incumbent party’s manifesto than the opposition, while the frequency of negative emotions will be highest in the opposition party’s manifesto than the incumbent
}}

A study conducted in campaign materials (including manifestos) across European elections indicates a correlation between the usage of emotions and the party's current status in government \citep{crab_2018}. To be exact, the party or parties who are part of the incumbent government increase the frequency of positive emotions across different campaign materials than the opposition. Regarding coalition governments, the party that controls the Prime Minister's position uses positive emotions more than its coalition partners \citep{crab_2018}. The reason incumbent parties, especially those that control the PM office, increasingly assert more positive language. This can be related to the voters directly assigning them greater responsibility as the position of Prime Minister is involved in the planning and execution of public policy, which can then attract mass public attention and even more scrutiny from the general public \citep{crab_2018}. This incentivizes the incumbent parties to improve the voter's perception of their government and its past decisions by positively portraying themselves (especially in times of national crisis or civil unrest).

However, in Finnish party programs, \citet{ohman_fin} did not find differences between the incumbent and opposition parties using emotion-associated words. Instead, they found the differences statistically insignificant except for the populist parties who used significantly more intense words to describe their party goals.

In terms of the opposition, due to its main role as a critic and a rival to the incumbents, it has been shown to exploit negative sentiments and language within its campaign materials as a strategy to undermine the incumbents' image \citep{louwerse_2021}. Based on these dynamics of a party's position in government and its impact on positive and negative language, the research formulates H1.

\textit{\textbf{Hypothesis 2 (H2):  Manifestos of ideologically similar parties will use positive language more frequently than negative language}}

\citet{kosmidis} found that in an election between parties that are similar in ideology and policy content (i.e., established centrist parties), they would compete using "emotional appeals." Specifically, these parties would primarily adopt positively associated words to captivate the same voter base \citep{kosmidis}. 

While the British Conservative Party and Labour Party have been recognized as ideological opposites from the 1900s to 1970s, the increased approval of policies under Thatcherism and Neo-liberalism starting from the 1980s in British politics have led to their gradual convergence in ideology and policy \citep{convergence}. This convergence has continued into the 21st century, with research finding that Labour Party and Conservative Party politicians continued to share neo-liberal values even after the Great Financial Crisis and succeeding economic crisis between 2008 and 2014 \citep{convergence}. Based on this, the research establishes H2 in which ideologically similar parties (the Conservative Party and the Labour Party) will have a higher frequency of positive language than negative language in their manifestos.


\textit{\textbf{Summary:}}
Both H1 and H2 aim to investigate how a party's government position (incumbent or opposition) and ideological positioning (similarities) will impact the type of emotions and their relative frequency in manifesto texts. 

\section{Data and Methods}

In the UK, manifestos are only released in preparation for the nation's general elections for the House of Commons, where the results of seats of MPs will determine which party (or parties) forms a government \citep{wikipedia_con}. Therefore, the data includes 12 general election manifestos, six each for the Conservative Party and Labour Party, published over six general elections (2001, 2005, 2010, 2015, 2017, and 2019). All manifestos have been accessed from two Wikipedia pages, each dedicated to both parties respectively, that present lists of available manifestos accessible by PDF files or redirected to websites with the manifesto texts (which were later converted and downloaded as PDF files) \citep{wikipedia_con, wikipedia_lab}.

The decision to retrieve manifestos from 2001 to the most recent election of 2019 was to provide enough data to observe any salient differences in the composition of emotive languages and sentiments when both parties either switched as the incumbent or opposition. From 2001 to 2010, the UK was led by a Labour government, with the Conservative Party as the primary opposition \citep{wikipedia_lab}. Their positions switched with the outcome of the 2010 election, where the Conservative Party formed a government still in power as of 2023 \citep{wikipedia_lab}. Therefore, 2001 to 2019 offers a sufficient timeline to pursue the research objective effectively.

For the text extraction from PDF files, a PDF text extractor called "pdfminer.six" was used for this research. Unlike other PDF extractors such as PyPDF2, pdfminer can compensate for two-column documents with relative success. After the extraction was successful, the NLTK sentence tokenizer \citep{bird2009natural} was used to separate the large string of text into a list of sentences, which were then processed, removing newlines and URL links and correcting Unicode formats.

\subsection{Method 1: VADER Sentiment Analysis}

This study used a popular lexicon-based sentiment analysis tool "VADER" by \citet{hutto2014vader} to investigate each manifesto's general composition of positive, negative, and neutral sentences within a manifesto text. 

VADER (Valence Aware Dictionary for Sentiment Reasoning) uses a lexicon-based approach to analyze each lexicon within texts and return an aggregated sentiment score of the text ranging from -1 to 1 (most negative and most positive respectively) \citep{calderon_2018}. To do so, VADER utilizes a dictionary that assigns sentiment scores to different lexical features and emoticons (i.e. emojis and colloquialisms). Moreover, VADER accounts for other text features, including capitalization, punctuation, and negation, making it a popular lexicon-based sentiment analysis tool, especially for social media texts \citep{calderon_2018}.

The manifestos were left as is (no stopword removal, lemmatization, etc., were implemented), and the texts (manifesto sentences) were processed by the VADER 'Sentiment Intensity Analyzer' object to calculate the compound score of the sentences. Each compound score of the sentences was translated into a readable format with a score of over 0.05 labeled as 'positive,' a score below -0.05 as 'negative,' and a score in between as 'neutral.' 

Within the contemporary field of Sentiment Analysis, lexicon-based methods have garnered a degree of criticism in regards to their general limitation in \textbf{"accuracy"} and \textbf{"validation"} as compared to data-driven methods. In prior Sentiment Analysis research, many found that lexicon-based methods' have lower efficacy when validating results with human annotators \citep{ohman-2021-validity, teodorescu2022frustratingly}. 

This phenomenon is well documented with several lexicon-based packages, such as R's Syuzhet \citep{jockers2017package}, which prioritizes word frequency to its analysis (leading to misclassification of sentiments to text) alongside their limited ability to detect negation. Nonetheless, studies have also shown that lexicon-based methods can still be a valid tool for analyzing the emotions and sentiments within texts. Research comparing the efficacy of VADER and TextBlob \cite{loria2018textblob} (another lexicon-based analysis tool) using tweets of COVID vaccines found that VADER classification generally corresponded better than TextBlob when directly comparing human annotation results, while another study analyzing tweets of the 2016 US election (exclusively using VADER) yielded satisfactory results of accuracy according to its authors \citep{georgios-alexandros_2022, elgagir_yang_2019}. Therefore, along with the benefit of low-cost implementation (i.e., the negation of classification by human annotators for machine learning models), lexicon-based methods, including VADER, are capable of providing useful and even desirable insights for researchers that are comparable to machine learning methods \citep{ohman-2021-validity}. Moreover, \citet{ohman-2021-validity} suggests that the lexicon-based approach has its merits when the purpose of the study is not to achieve higher numerical accuracy in sentiment classification but to focus on investigating emotion-associated words.

Therefore, the application of VADER is a useful and justified method in this study, as the goal is to provide a generalized overview of positive and negative contents (via sentences) within the manifestos and attempt to understand the trend of emotion usage while being able to maintain a satisfying level of accuracy compared to other similar lexicon-based packages.

\subsection{Method 2: NRC Emotion lexicons}

Along with VADER, NRC Word-Emotion Associations (EmoLex), another lexicon-based emotion detection and analysis method was used. EmoLex was created via crowd-sourcing to associate existing English lexicons with various emotions by manual human annotation \citep{mohammad_turney_2012}. Unlike VADER, which is only capable of ternary sentiment classifications, EmoLex annotators have additionally assigned each lexicon with the eight basic affect categories: joy, trust, anticipation, surprise, fear, sadness, anger, and disgust \citep{mohammad_turney_2012}. In effect, EmoLex is capable of analyzing the deeper dynamics of specific affect emotion-laden lexicons and their distribution beyond the limited binary positive and negative sentiment classification, a task considered the most suitable for lexicon-based methods and incorporated into a previous study on post-war Finnish manifestos \citep{ohman-2021-validity, ohman_fin}. Furthermore, \citet{ohman-2021-validity} proposes that emotion or sentiment analysis research using lexicon-based models should be evaluated by sanity checks to verify the usefulness and accuracy of any given results. As such, EmoLex also aims to act as an evaluation of the VADER results (and vice versa) to check the rationality and validity of the results obtained. 

While the deployment of EmoLex was largely similar to that of VADER, we further included the lemmatization of sentences and the removal of punctuation and other non-alphanumeric characters. The processed sentences were later merged as a single text used for EmoLex analysis. In this study, the affect frequency, which returns the frequency with the range value of 0 to 1 for the two sentiments and eight affect emotions from the analyzed manifesto text, was used to visualize the emotion-word frequencies.

\section{Results}
\subsection{VADER Sentiment Analysis Results}

The results of VADER sentiment analysis of both parties' manifestos are shown in table \ref{labour_res} and table \ref{Cons_sent}, respectively.  

\begin{table}[htbp!]
    \newcolumntype{+}{>{\global\let\currentrowstyle\relax}}
    \newcolumntype{^}{>{\currentrowstyle}}
    \newcommand{\rowstyle}[1]{\gdef\currentrowstyle{#1}%
        #1\ignorespaces
    }
    \centering
    \resizebox{\textwidth}{!}{%
    \begin{tabular}{+>{\bfseries}l^c^c^c^c^c^c^c}
        \hline
        \rowstyle{\bfseries}
        Year & Sentences & GOV Status & Pos\_share(\%) & Pos Change & Neg\_share(\%) & Neg Change & Neut\_share(\%) \\
        2001          & 977                & Incumbent           & 62.641                  & 0.0                 & 16.07                   & 0.0                 & 21.29                    \\
        2005          & 801                & Incumbent           & 63.92                   & 1.28                & 19.725                  & 3.7                 & 16.355                   \\
        2010          & 1313               & Incumbent           & 66.565                  & 2.64                & 16.375                  & -3.4                & 17.06                    \\
        2015          & 865                & Opposition          & 60.231                  & -6.33               & 22.89                   & 6.5                 & 16.879                   \\
        2017          & 1136               & Opposition          & 53.257                  & -6.97               & 24.032                  & 1.1                 & 22.711                   \\
        2019          & 1192               & Opposition          & 50.336                  & -2.92               & 30.369                  & 6.3                 & 19.295                   \\
        \hline
    \end{tabular}
    }
    \caption{Sentiment Results of Labour Party Manifesto sentences (2001 - 2019)}
    \label{labour_res}
\end{table}

\begin{table}[htbp!]
    \newcolumntype{+}{>{\global\let\currentrowstyle\relax}}
    \newcolumntype{^}{>{\currentrowstyle}}
    \newcommand{\rowstyle}[1]{\gdef\currentrowstyle{#1}%
        #1\ignorespaces
    }
    \centering
    \resizebox{\textwidth}{!}{%
    \begin{tabular}{+>{\bfseries}l^c^c^c^c^c^c^c}
        \hline
        \rowstyle{\bfseries}
        Year & Sentences & GOV Status & Pos\_share(\%) & Pos Change & Neg\_share(\%) & Neg Change & Neut\_share(\%) \\
        2001          & 680                & Opposition          & 48.382                  & 0.0                 & 25.735                  & 0.0                 & 25.882                   \\
        2005          & 415                & Opposition          & 53.012                  & 4.63                & 20.241                  & -5.5                & 26.747                   \\
        2010          & 494                & Opposition          & 58.3                    & 5.29                & 22.065                  & 1.8                 & 19.636                   \\
        2015          & 283                & Incumbent           & 75.618                  & 17.32               & 13.428                  & -8.6                & 10.954                   \\
        2017          & 1275               & Incumbent           & 70.667                  & -4.95               & 13.961                  & 0.5                 & 15.373                   \\
        2019          & 974                & Incumbent           & 64.682                  & -5.98               & 15.811                  & 1.8                 & 19.507                   \\
        \hline
    \end{tabular}
    }
    \caption{Sentiment Results of Conservative Party Manifesto sentences (2001 - 2019)}
    \label{Cons_sent}
\end{table}

The results in table \ref{Cons_sent} for the Labour Party show that sentences with positive sentiments compromised over 60\% of the total share of sentences within the manifestos. This is seen in the Pos\_share (percentage share of positive sentences out of the total) during its incumbency, with its Pos Change (the change in shares of positive sentences from the previous election) showing a small increase in share from 2001 to 2010. These 3 (2001, 2005, and 2010) Pos\_share figures were each higher than those of Conservative Party Manifestos published during the corresponding election year as presented in table \ref{Cons_sent}. However, after the 2010 election, when the Labour Party was relegated to the opposition, there appears to be a gradual drop in Labour Party's Pos\_share from 2015 to 2019, with the Pos\_share of the Labour Party's last manifesto in 2019 being only around 50\%. At the same time, post-2010 Labour manifestos saw an increase in total shares of negative sentences (Neg\_share), further seen by the highest Neg Change (the change in total shares of negative sentences from the previous election) of over 6 points in 2015 and 2019. This has led the Neg\_share of the Labour Party's 2019 manifesto to be 30\%, doubling the Neg\_share of the Labour Party's 2001 manifesto. Additionally, as the opposition, the Neg\_share of all Labour Party manifestos in the last three elections (2015, 2017, and 2019) was larger than all the Conservative Party manifestos published in the same election cycle. 


The results of the Conservative Party manifestos are presented in table \ref{Cons_sent}. As mentioned before, during their time as the opposition from 2001 to 2010, the Pos\_share of the Conservative Party's manifesto was always smaller than those of the Labour Party manifestos. This is not the case for Neg\_share, which, for the first three elections of 2001, 2005, and 2010, the Conservative Party manifestos had a higher share of negative sentences than the Labour Party. This trend changed after the Conservative Party returned to power in the 2010 elections, becoming the incumbent. Starting from the 2015 election, the Conservative Party's Pos\_share immediately overtook the Labour Party. Specifically, the share of positive sentences in the Conservative Party's 2015 manifesto saw a 17-point increase in Pos Change from 2010 and a 15-point lead against Labour's manifesto published in the same year. While their Pos Change in 2017 and 2019 turned negative (indicating the decline of total shares in positive sentences in Conservative manifestos), the Conservative Party's Pos\_share were much higher than the Labour Party in the last two elections. 


\subsection{EmoLex Emotion Analysis Results}

Moving to the EmoLex results, Appendix 1 presents the 6 line graphs representing the frequencies of Positive/Negative sentiments and the 8 affect emotions via TJA and FASD graphs for the Labour Party and the Conservative Party manifestos. (Note: The TJA Graph represents positively associated emotions: Trust, Joy, and Anticipation. The FASD Graph represents negatively associated emotions: Fear, Anger, Sadness, and Disgust)

Starting with the Labour Party's graph for the frequency of Positive and Negative lexicons, during the 2001-2010 period, positive lexicons had a frequency of 0.3 while negative lexicons were just 0.1. Comparing this to the graph for the Conservative Party, we find that the frequency of the Labour Party for positive lexicons was higher in the first three elections during this period. In return, however, the Conservative Party had a higher frequency in the negative lexicon. This pattern, however, isn't reflected in the TJA and FASD Graphs. For example, while the Labour Party had a lead in the frequency of Trust lexicons in the TJA graph against the Conservative Party in the 2001 election, the Conservative Party had overtaken them in the following election in 2005 despite being the opposition. Moreover, neither party had a constant lead in the frequency of fear, anger, and sadness during this period.

Since the 2010 elections, however, a salient pattern has emerged where the Labour Party manifestos began showing an increase, eventually taking the lead in the frequency of negative and negatively associated FASD lexicons in the 2019 election. Specifically, the frequency of negative lexicons in the Labour Party's 2010 manifesto was initially 0.1; this increased to around 0.14 by 2019. Regarding the lexicons related to fear, it also saw an increase starting from 0.05 in 2010 to below 0.08 at the same time by 2019. In comparison, the frequency of negative and FASD lexicons within the Conservative Party manifestos shows a minor decline over time. In its replacement, positive and positively associated TJA lexicons are more frequently used in Conservative Party manifestos over time. And by the 2019 election, the Conservative Party used positive and positively associated TJA lexicons more frequently than the Labour Party. 
\pagebreak

\begin{figure}[htbp!]
    \centering
    \includegraphics[width=8.5cm]{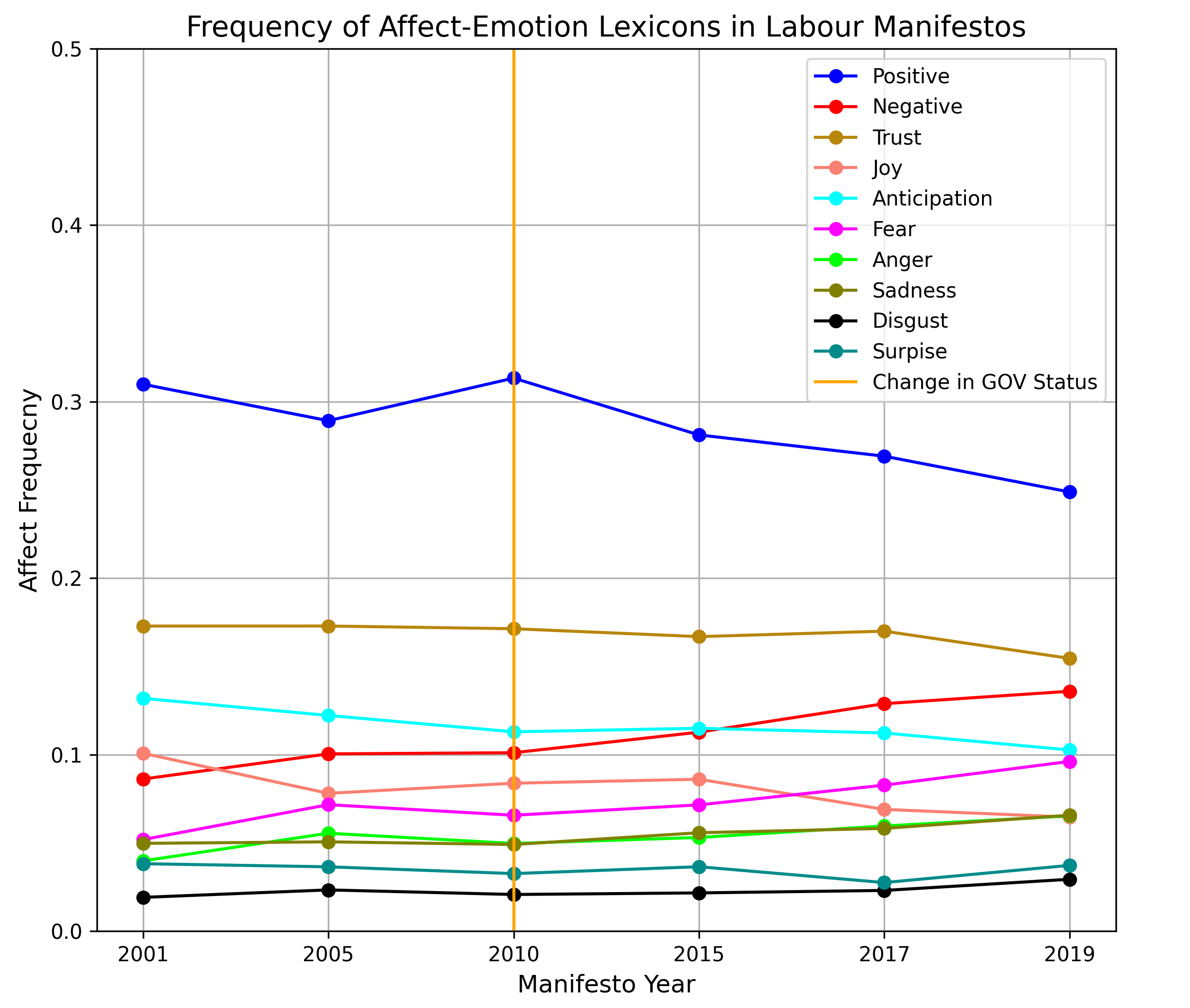}
    \caption{Labour party change in emotion-associated content}
    \label{fig:lab}
\end{figure}

\begin{figure}[htbp!]
    \centering
    \includegraphics[width=8.5cm]{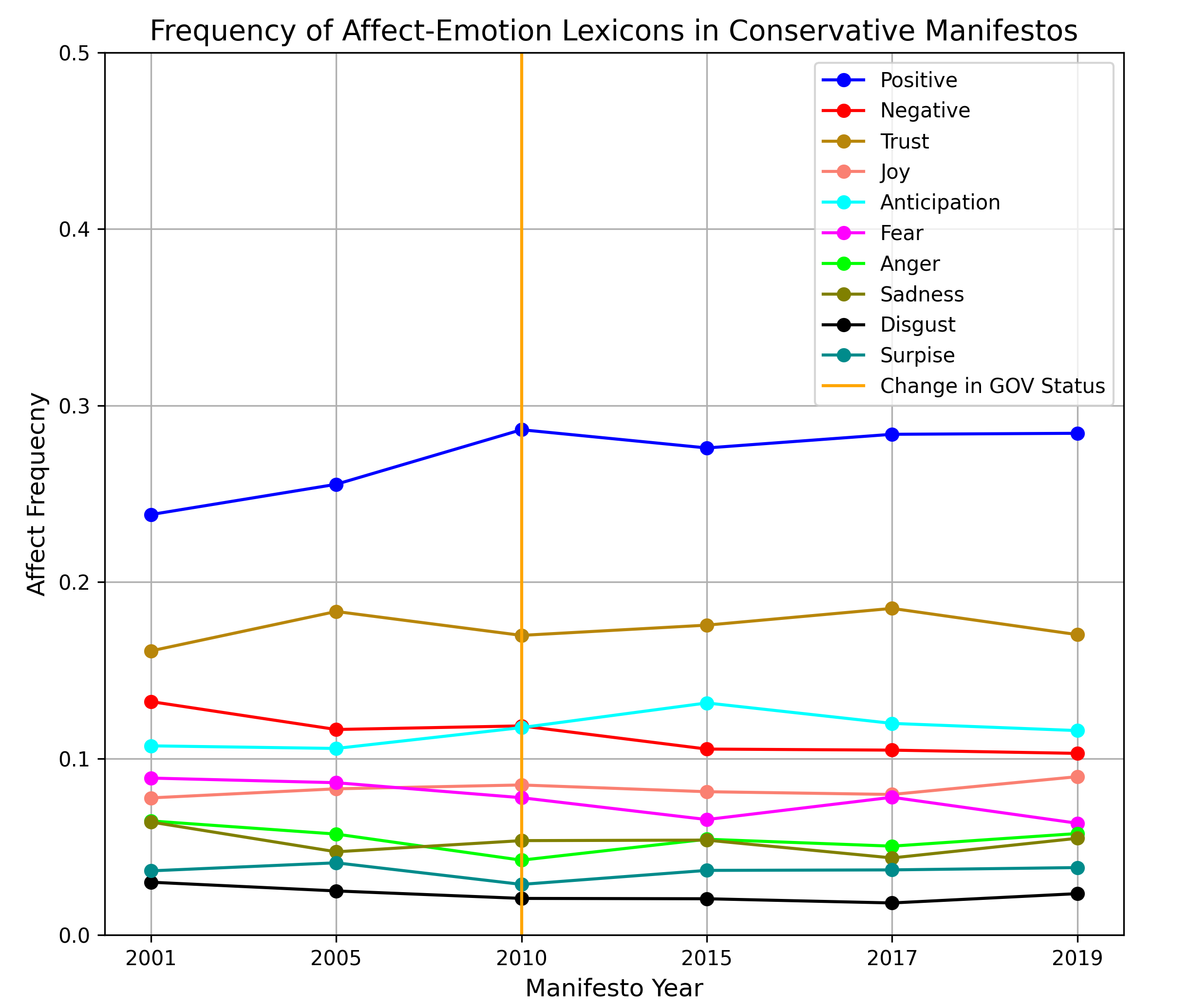}
    \caption{Conservative party change in emotion-associated content}
    \label{fig:con}
\end{figure}

Figure \ref{fig:lab} and Figure \ref{fig:con}, represent the combined result of the 3 line graphs given to each party. 

Combining all the EmoLex results into a single graph (Figure \ref{fig:lab} for the Labour Party and Figure \ref{fig:con} for the Conservative Party), we find that positive languages (especially positive sentiments and the emotion of trust) are most frequently adopted for both parties' manifestos regardless of their position as opposition or incumbent. Nevertheless, there is a notable trend in the Labour manifestos in Figure \ref{fig:lab} where negative emotions gradually replace the share of frequency of certain positive emotions. This is particularly highlighted in the 2019 manifesto, where the frequency of negative lexicons was closely behind trust lexicons for the Labour Party.

\begin{figure}[hbtp!]
    \centering
    \includegraphics[width=11cm]{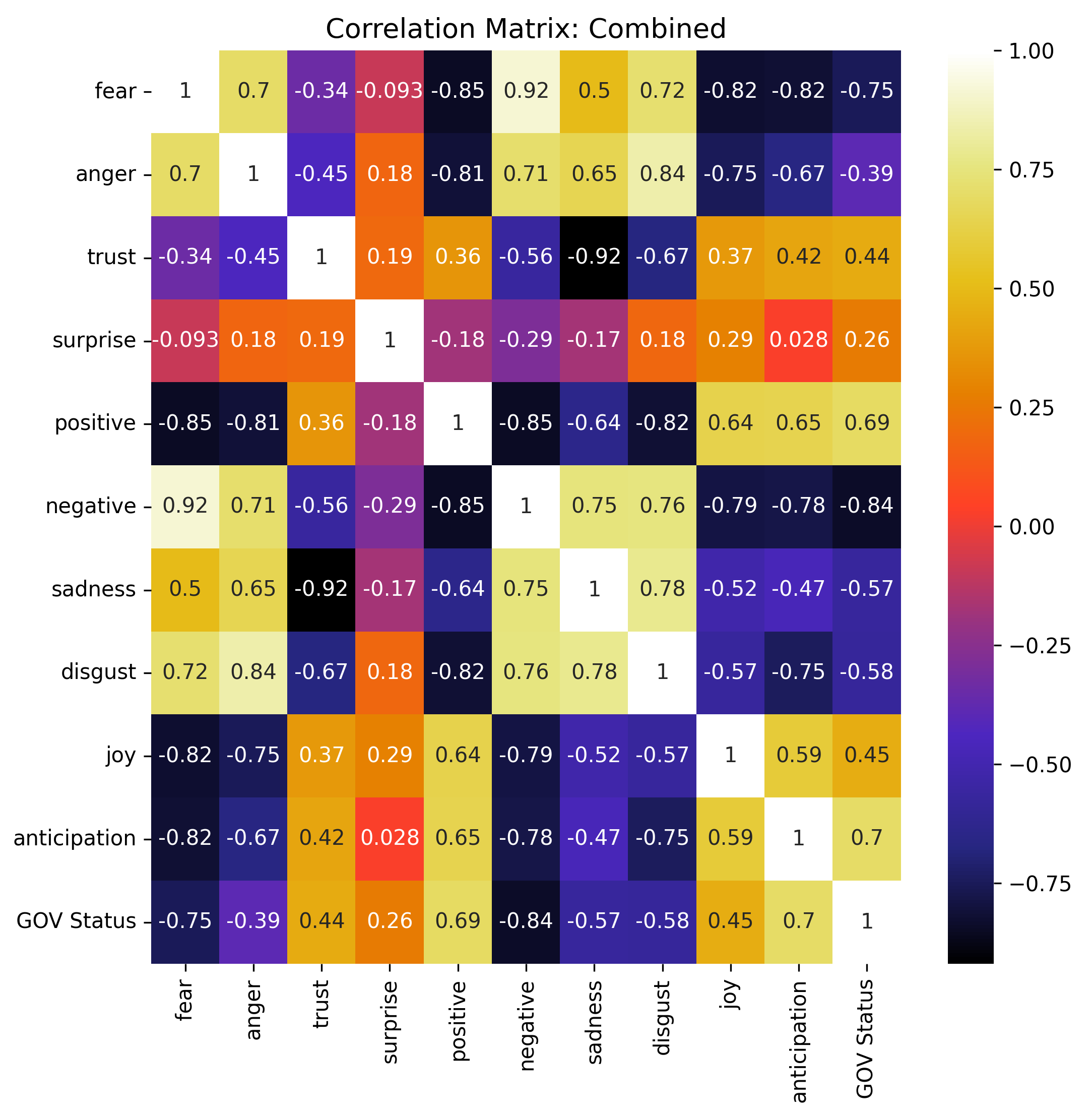}
    \caption{Correlation matrix for both parties combined}
    \label{corr}
\end{figure}

Figure \ref{corr} is a heat map created from both parties' combined EmoLex results to identify the correlation between the frequency of sentiment and affect emotions against the party's status in government (as incumbent or opposition). The results indicate that positive emotions are relatively correlated with the party's status in government (GOV Status); where a party becomes the incumbent, positive language will likely increase. On the other hand, negative emotions are negatively correlated, indicating that they will increase when the party becomes the opposition. 

The results summarized in this section will be further analyzed with the hypotheses addressed in section 2.

\section{Analysis of Results}

\textit{\textbf{Assessment of H1: The frequency of positive emotions will be highest in the incumbent party’s manifesto than the opposition, while the frequency of negative emotions will be highest in the opposition party’s manifesto than the incumbent
}}

The results presented in the VADER sentiment analysis and EmoLex effectively support the assumptions of the first hypothesis (H1). This is presented in figure \ref{corr}, where we find a strong correlation between government status and positive/negative lexicons. The findings are further supported by the VADER results exhibited in table \ref{fig:lab} and table \ref{fig:con}. Based on these results, the incumbent Conservative Party had more total shares of positive sentences in all post-2010 elections than the opposition Labour Party. However, as the opposition, the Labour Party had more shares of negative sentences than the Conservative Party. The pattern presented in the results from VADER is also reciprocated in the EmoLex results in the post-2010 elections. According to Appendix 1 and figure \ref{fig:lab}, after being relegated to the opposition, the Labour Party had increased the frequency of negative lexicons in their manifestos, eventually surpassing the Conservative Party. During the same period, the Conservative Party, now the incumbent, began increasing the frequency of positive lexicons and later surpassing the Labour Party manifestos as demonstrated in figure \ref{fig:con}.

Both findings for VADER and EmoLex are in line with previous research, in which incumbent parties, due to their increased responsibility of governance associated by the public, would use positive language at a higher rate to improve their image towards the public \citep{crab_2018}. At the same time, the opposition party would use more negative language to criticize and tarnish the incumbent(s), such as pointing to their failure of governance. This pattern arguably manifested in UK general election manifestos of 2017 and 2019 following the polarizing 2016 Brexit referendum. Given that the Conservative Party was largely held responsible by the British public to guide the country after the Brexit decision, their 2017 and 2019 manifesto needed to include positively reinforcing messages to boost public confidence in their government \citep{Bonnet_2021}. In contrast, the Labour Party used its status as the opposition to scrutinize the Conservative government, including their 'mishandling' of Brexit and critically opposing their public policies \citep{hayton_2021}. These dynamics have arguably led the Conservative manifestos to increase and maintain a high frequency of positive emotions within their manifestos. In contrast, the Labour Party manifestos displayed a higher frequency of negative emotions after the 2010 election, validating H1.

\textit{\textbf{Assessment of H2: Manifestos of ideologically similar parties will use positive language more frequently than negative language}}

The VADER and EmoLex results also show that positive languages had the highest frequency in both parties across all elections, regardless of the party's position as incumbent or opposition. This supports H2, where parties with ideological similarities would largely compete with positive language in their manifestos. The Pos\_share of table \ref{fig:lab} and table \ref{fig:con} show that sentences with positive sentiments constitute at least half of the total sentences within the manifesto. In contrast, Neg\_share rarely constituted a third of the total. Additionally, figure \ref{fig:lab} and figure \ref{corr} show that positive lexicons (excluding TJA lexicons) accounted for the highest frequency, as predicted in previous research relating to H2.

Despite the results, when comparing the Conservative Party's EmoLex frequencies, the Labour Party resorted to increasing the frequency of negative lexicons (including FASD emotions) and even displacing several positively associated TJA lexicons in frequency by 2019. While these trends can partially be explained through the analysis of H1, another factor that may have influenced Labour Party manifestos to increase the usage of negative lexicons, especially in 2017 and 2019, is also the left-wing shift of the Labour Party under the leadership of Jeremy Corbyn from 2015 to 2020 \citep{corbyn_radical}. Under Corbyn, many had remarked on the Labour Party's increasing return to advocating traditional left-wing economic and social policies, thereby shifting the party's ideology to the left and away from the past centrist positions, leading some to describe this move as 'radicalization' of the party. While some question the extent of the Labour Party's ideological shifting under Corbyn's leadership, close research on policies presented in Labour 2017 and 2019 manifesto demonstrates a clear trend of left-leaning shift \citep{corbyn_radical, jacobs_hindmoor_2022}.

Referring to the study by \citet{kosmidis}, parties with similar ideologies and policy positions often rely on positive "emotional appeals" during their campaign to compete over the same electorate. Radical fringe parties, however, would commonly adopt negative languages to mobilize disaffected voters, as suggested by \citet{crab_2018}. This may help to explain why negative and FASD lexicons overtake certain TJA lexicons in the Labour Party's manifesto from 2017 to 2019, as seen in Figure 1, as it likely represents the growing radicalization of the Labour Party, which increasingly promoted radical policies compared to the Conservative Party, impacting their amount of positive lexicons as explained by \citet{crab_2018}.

On the surface, results for VADER and EmoLex support H2, with positive languages holding the highest frequency through all elections in both the Labour Party and the Conservative Party regardless of their position as incumbent or opposition. However, the critical rise in the frequency of negative emotions in the Labour Party's 2017 and 2019 manifesto likely reflects its political radicalization and partial ideological divergence from the Conservative Party. Nonetheless, the results can be observed as evidence of how ideological positioning can factor in the dynamics of emotional language within manifestos.

\section{Conclusion}

The primary aim of this research was to investigate the impact of a party's current government status and position of ideology on the use of emotional language in their election manifesto. Therefore, using the recent UK election manifestos as data, we tested the two hypotheses to investigate this relationship. To summarize the findings, VADER and EmoLex analysis underscores the strong influence a party's incumbent or opposition status and ideological positioning have on the frequency of positive and negative language in manifestos. 

In future work, we hope to address several limitations, including preprocessing of texts, adding domain-specific terms to the emotion lexicons, and incorporating additional manifesto data. As it currently stands, the vastly different formats of the manifestos might have introduced some artefacts in the text extraction process, of which the majority were manually removed. However, some may remain.

Additionally, since the EmoLex was originally annotated using North American annotators, some cultural differences in emotion associations might exist. Furthermore, the domain of the manifestos is very specific, and likely, several terms were not properly included. In future studies, the lexicon should be made more domain-specific.

Finally, as the manifesto data is limited to those produced since the 2000s, it only presents a period when both the Labour Party and the Conservative Party already displayed strong similarities in ideology. While H2 aimed to investigate the impact this had on the emotive frequency in manifestos, future studies should incorporate additional data from previous elections when both parties presented far stronger ideological and policy distinctions to further test the validity of our hypotheses.

\bibliographystyle{plainnat}
\bibliography{References}
\pagebreak
\appendix\footnotesize
\section{Appendix 1: Figures for Emolex}
\begin{figure}[hbtp]
    \includegraphics[scale=0.14]{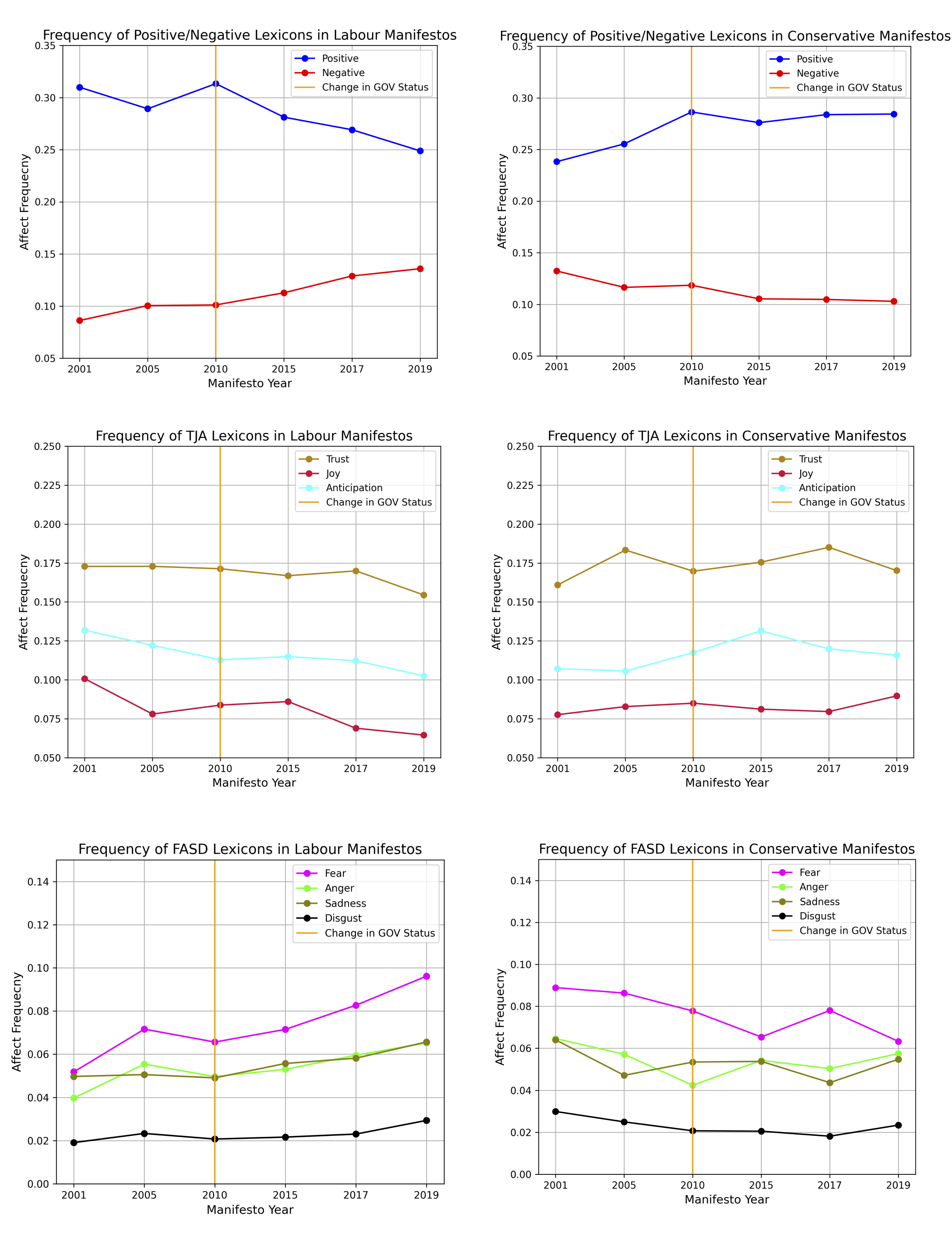}
    \centering
\end{figure}
\end{document}